\documentclass{article}
\usepackage[utf8]{inputenc}
\usepackage{graphicx}
\usepackage{mathrsfs} 
\usepackage{algorithm}
\usepackage{multirow}
\usepackage{subfigure}
\usepackage{pgfplots}
\usepgfplotslibrary{groupplots}
\pgfplotsset{compat=1.12}
\usepackage{tikz}
\usepackage{tikzscale}
\usepackage{algorithm}
\usepackage[noend]{algpseudocode}
\usepackage{amsmath}
\usepackage{stfloats}
\usepackage[a4paper,margin=1in,footskip=0.25in]{geometry}

\title{Knowledge driven Description Synthesis for Floor Plan Interpretation}
\author{Shreya Goyal, Chiranjoy Chattopadhyay, and Gaurav Bhatnagar}
\date{March 2021}

\begin{document}

\maketitle

\begin{abstract}
Image captioning is a widely known problem in the area of AI. Caption generation from floor plan images has applications in indoor path planning, real estate, and providing architectural solutions. Several methods have been explored in literature for generating captions or semi-structured descriptions from floor plan images. Since only the caption is insufficient to capture fine-grained details, researchers also proposed descriptive paragraphs from images. However, these descriptions have a rigid structure and lack flexibility, making it difficult to use them in real-time scenarios. This paper offers two models, Description Synthesis from Image Cue (DSIC) and Transformer Based Description Generation (TBDG), for the floor plan image to text generation to fill the gaps in existing methods. These two models take advantage of modern deep neural networks for visual feature extraction and text generation. The difference between both models is in the way they take input from the floor plan image. The DSIC model takes only visual features automatically extracted by a deep neural network, while the TBDG model learns textual captions extracted from input floor plan images with paragraphs. The specific keywords generated in TBDG and understanding them with paragraphs make it more robust in a general floor plan image. Experiments were carried out on a large scale publicly available dataset and compared with state-of-the-art techniques to show the proposed model’s superiority. 
\end{abstract}

\section{Introduction}
In document image analysis, a floor plan is a graphical document that aids architects in showing the interior of a building. Floor plan image analysis involves semantic segmentation, symbol spotting, and identifying a relationship between them. Describing a floor plan in natural language is a task that has applications in robotics, real-estate business, and automation. However, there are several challenges when it comes to narrating a graphical document in natural language. A graphical document is not similar to a natural photograph that has an essential feature in every pixel. Hence, traditional approaches using image features with textual description fails in this context. The graphical document requires specific information for their description to make it more meaningful. Hence, cues taken directly from an images are not very efficient in this context. There are several approaches available for language modeling and text generation in which the encoder-decoder framework is the most popular choice. In the image to text generation, CNN-RNN (CNN acting as an encoder, RNN as a decoder) is widely used in literature. The variants of RNN is varied in the decoder as LSTM, Bi-LSTM, and GRU.

\begin{figure}[t]
		\centering
		\includegraphics[scale=.37]{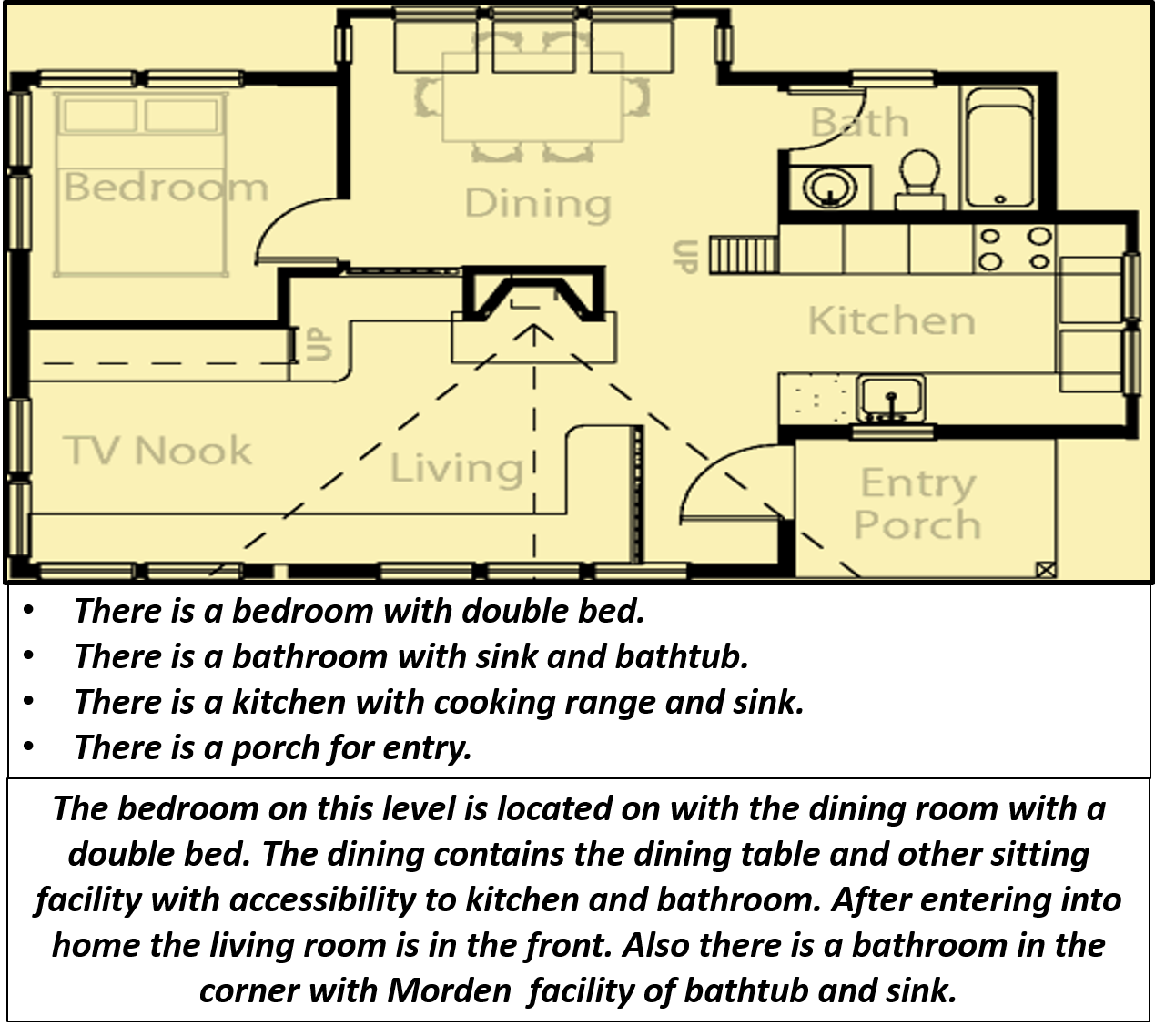}
		\caption{An illustration of the he proposed problem domain with the desired output.}
		\label{fig:prob}
\end{figure}

\begin{figure*}[!b]
		\centering
		\includegraphics[width= \linewidth]{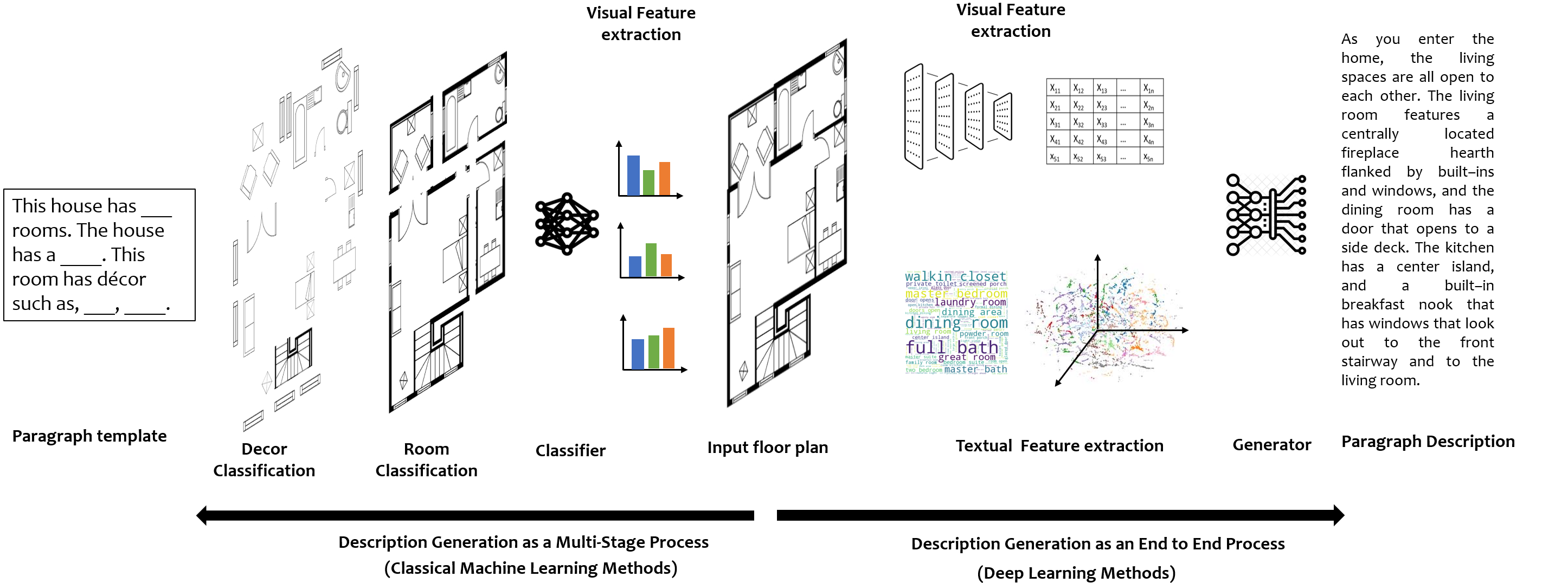}
		\caption{An illustration of the proposed work of generating textual description from floor plans.}
		\label{fig:work_idea}
\end{figure*}

Figure \ref{fig:prob} depicts the proposed problem with the desired output. The generated captions in Fig. \ref{fig:prob} (second row from the top) are very structured and contain limited information. The bottom row provides more realistic descriptions. We take advantage of both image cues and the word signals to generate specific and meaningful descriptions from the floor plan images. The proposed work is an extension of the paper BRIDGE \cite{bridge2019}, where a large scale floor plan image dataset and annotations were proposed. This paper’s proposed work leverages those annotations by offering multi-sentence paragraph generation solutions from floor plan images.

Figure \ref{fig:work_idea} depicts the overall flow of our proposed method. We extend the idea of extracting information from floor plan images in a multi-staged pipeline using classical machine learning methods and generating textual descriptions by using semi-structured sentence based models proposed in the literature. This direction’s previous work is extended by offering models that learn textual features with visual features in an end to end framework.
We propose two models, Description Synthesis from Image Cue (DSIC) and Transformer Based Description Generation (TBDG), where TBDG is more robust than DSIC. These two proposed models differ in the way the decoder receives the input. In DSIC, region-wise visual features are learned with textual features, and a paragraph description is generated. In contrast, TBDG learns region-wise captions with region wise features, and those text features are given as input to the decoder model to create a paragraph. We further propose a deep learning based multi-staged pipeline for description generation in order to prove the superiority of end-to-end learning models on multi-staged pipelines.

\textbf{Uniqueness of the proposed work:} In the previous work, \cite{goyal2018asysst,goyal2019sugaman,goyal2018plan2text}, only visual elements are learned and classified in a multi-staged manner using classical machine learning approaches. Tasks such as semantic segmentation, room classification, and decor classification are performed in a sequential pipeline using classical machine learning methods. In \cite{bridge2019}, the similar visual elements are learned and classified in part by part manner using a deep neural network. In contrast with the existing approaches, in this paper all the visual information from floor plan images and textual features are learned together in an end to end deep learning framework, and a holistic description for the same is generated.

\textbf{Organization of the paper:} 
The rest of the paper is organized in the following way. Section \ref{sec:litsurv} highlights the related works. Section \ref{sec:DSIC} and \ref{sec:TBDG} describes the two proposed models. Section \ref{sec:exp} describes the experimental setup and the evaluation metrics followed for performance analysis. Section \ref{sec:baseline} discusses the results generated using proposed models. Section \ref{sec:des_gen} describes the comparative analysis of various stages involved in description generation models and their qualitative and quantitative comparison, while the paper is concluded in Sec. \ref{sec:conclusion}.  

\section{Related work}
\label{sec:litsurv}
\subsection{Publicly available Floor plan Datasets}
In the literature, the publicly available datasets are: ROBIN \cite{sharma2017daniel}, CVC-FP \cite{de2015cvc}, SESYD \cite{delalandre2010generation}, BRIDGE \cite{bridge2019}, and FPLAN-POLY \cite{barducci2012object}. However, apart from the BRIDGE, the others contain very few sample images and do not contain annotations for objects and their textual descriptions. These datasets were proposed for segmentation, retrieval, and layout analysis, which are not suitable for caption generation and description synthesis tasks. With the advent of deep neural networks, tasks such as symbol spotting, caption generation, retrieval, semantic segmentation are getting more accurate and robust. To meet the requirement of a large number of samples and corresponding annotations, the  BRIDGE \cite{bridge2019} dataset was proposed, having $13000$-floor plan images, along with annotations. There is high variability in the way decor symbols has been represented across these datasets. To overcome this limitation, samples from \cite{delalandre2010generation,de2015cvc,sharma2017daniel} has been included in the BRIDGE dataset, and decor symbol annotations are done. Hence, a large portion of variable decor symbols has been covered in BRIDGE dataset.

There are many large scale datasets publicly available in the literature in the context of natural images \cite{krishna2017visual,lin2014microsoft}, which has many realistic images along with their descriptions or captions annotations, region graphs, and other metadata. For example, \cite{krishna2017visual} connects $108,077$ images with textual annotation, $5.4$ Million region descriptions, and other annotations for various tasks such as caption generations, visual question answering. MS-COCO and MS-COCO captions \cite{chen2015microsoft} are examples of datasets that contain over $330000$ images and over one and a half million captions ($5$ captions per image).

\subsection{Object detection and classification}

Researchers have explored handcrafted features and conventional machine learning models \cite{dutta2013symbol,dutta2011symbol,viola2001rapid,qureshi2007spotting,adam2000symbol} for the symbol spotting task in document images. As the deep neural network models are getting popular, methods like YOLO \cite{redmon2016you}, Fast-RCNN \cite{girshick2015fast}, Faster-RCNN \cite{ren2015faster} in the context of natural images were proposed. All the YOLO family-based algorithms, \cite{redmon2016you,redmon2017yolo9000,redmon2018yolov3} are region classification based methods, which is a single neural network trained end to end and predicts bounding boxes and class labels directly. However, all the R-CNN family-based models \cite{girshick2015fast,ren2015faster,he2017mask,he2015spatial}, are region proposal based methods, which extracts several regions from the input image and extract features from those regions using CNN and classify them using a classifier. In one of the work \cite{ziran2018object}, symbol spotting in floor plans using YOLO and Fast-RCNN has also been explored. In the same line \cite{rezvanifar2020symbol} has performed symbol spotting in floor plans using the YOLO network. Apart from floor plans, symbol spotting or object detection in document images task has been explored by several researchers. Examples include detecting tables, equations, figures \cite{yi2017cnn,saha2019graphical,schreiber2017deepdesrt}, signatures and logos \cite{sharma2018signature,su2020scalable}. In \cite{khan2020comparative}, a comparative study of recognition of symbol spotting on graphical documents were presented. 

\subsection{Image description generation}

Image description generation is a challenging task in AI. Template-based retrievals, n-grams, grammar rules, RNN, LSTM, GRU, are some example approaches to solve the problem. These methods work with image modality features by extracting information related to image using conventional architectures. Some of the initial work in this direction, \cite{farhadi2010every,kulkarni2013babytalk,li2011composing,ordonez2011im2text} have used computer vision methods for extracting attributes from an image and generated sentences using retrieval and n-gram model. 

Johnson et al. \cite{johnson2016densecap} proposed an algorithm to generate region-wise captions. The Hierarchical recurrent network \cite{krause2017hierarchical} uses two RNNs to generate paragraphs from an image. In \cite{wang2018cnn+}, two CNN networks are used, where one of them is used as an encoder for image features, and the other is used as a decoder for language generation. Similarly \cite{chatterjee2018diverse} has used a sentence topic generator network from visual features of images, and RNN is used for sentence generation. In \cite{wang2018look}, depth aware attention model is used for generating a detailed paragraph from an image. In \cite{mao2018show}, Latent Dirichlet allocation (LDA) is used to mine topics of interest from textual descriptions and developed multiple topic-oriented sentences for image description. In \cite{yao2017boosting}, CNN, and the RNN framework is used to capture visual features and generate descriptions. The image description is also generated from a stream of images in \cite{park2015expressing} by using CNN and RNN for a visual feature and sequence encoding and retrieving sentences from an existing database. A similar line \cite{liu2017let} description has been generated as storytelling from images by using the Bi-directional attention-based RNN model.

\subsection{Language modelling}

Now a days, since, deep neural networks are very successful in natural language processing, learning text for generating description using sequence to sequence models are natural choice. In the work \cite{sutskever2014sequence} has proposed seq2seq learning model for learning and modelling language by LSTM models. Also, \cite{bahdanau2014neural} and \cite{luong2015effective} are the initial models which modelled language by aligning input sequence to a target sequence using attention based models. The neural machine translation models has also been used in text summarization tasks such as \cite{rush2015neural} and \cite{nallapati2016abstractive}. In the next section the proposed models for description generation from floor plan images are described in details. 

\section{Description Synthesis from Image Cue (DSIC)}
\label{sec:DSIC}
We have described floor plan images in the proposed model by extracting region-wise visual features from image and learning paragraphs by providing them to a decoder network. The region proposal network (RPN) act as the encoder, and a hierarchical RNN structure act as the decoder. The system is trained in an end to end manner. We describe each step in detail next.

\subsection{Visual feature extraction} 
\label{sec:hrnn}

We adopt a hierarchical RNN based approach as a decoder framework. Figure \ref{fig:hrnn} depicts a typical architecture of the proposed model. The dataset contains image $(I)$ and its corresponding paragraph description $(K)$ in the proposed approach. The CNN is used along with a RPN to generate region proposals, $R_1, R_2,...,R_n$. We extracted the top $5$ region proposals for this approach and pooled them in a single pooling vector $P$ using a projection matrix. In DSIC, two RNNs are used in a hierarchy, where one is used for learning sentence topic vectors from pooled features, and the other is used for learning words for respective sentence topic vector. In DSIC, the top $5$ regions are extracted because there are average $5$ sentences per paragraph in \cite{bridge2019}.

\subsection{Region Pooling:} All the extracted regions $R_i$ are pooled in a vector $P$ by taking the projection of each region vector $R_i$ with a projection matrix $M$ and taking an element wise maximum. Dimension of the pooled vector is same as the region vectors and defined as $P= \max_{i=1}^n{(MR_i+bias)}$
The projection matrix is trained end to end with the sentence RNN and the word RNN. The pooled vector $P$, compactly represents all the regions $R_i$s. 

\subsection{Hierarchical RNN structure:} 
\begin{figure}[t]
		\centering
		\includegraphics[scale=.35]{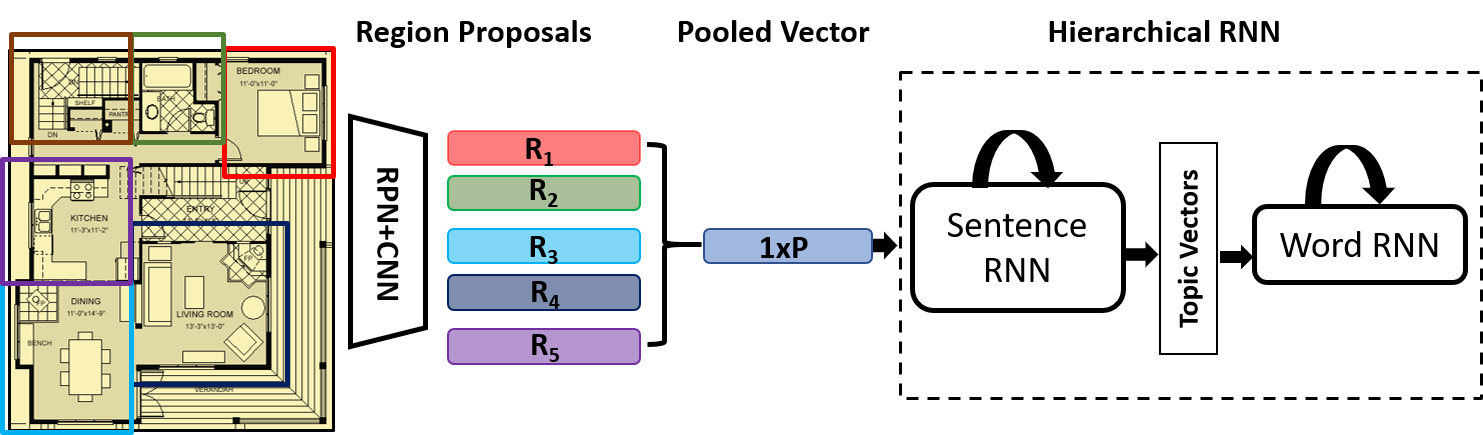}
		\caption{Hierarchical RNN to yield paragraph from floor plans.}
		\label{fig:hrnn}
\end{figure}

This network, as shown in Fig. \ref{fig:hrnn}, contains two units of the RNN network. One is sentence level (S-RNN) and the other is word-level (W-RNN). The S-RNN is a single-layered, used for generating sentence topic vector for each sentence, and decides the number of sentences to be generated. W-RNN is a two-layered and takes the sentence topic vectors as input and generates words in each sentence. Instead of using one single RNN as a decoder, which would have to regress over a long sequence of words and make training the language model harder, two RNN networks are taken in a hierarchy. The choice of networks for both RNNs is kept as LSTM networks since they can learn long-term dependencies than a vanilla RNN. The S-RNN is followed by $2$ layered fully connected network, which generates a topic vector to be given as input to W-RNN after processing the hidden states from RNN. The W-RNN takes topic vector and word level embeddings for the respective sentence as input. A probability distribution is generated for each word in the vocabulary, where is threshold, $Th$ is taken as $0.5$, which generates further words for each sentence. 

\subsection{Training:} 
At this stage, pooled vector $P_i$ generated from region proposals are taken as input to the sentence level RNN for each image $I$ and respective paragraph $K$. Each input maximum of $5$ sentences and $60$ words are generated (empirically identified based upon validation performance). Hence at each stage, $Sent_{max}=5$ copies of word RNN and topic vector is generated by the sentence RNN for each word RNN for, $Word_{max}$=$60$ timestamps. 

\begin{multline} \label{eq:hrnnloss}
    loss(I,K)= \beta_{sent}*\sum_{i=1}^{Sent_{max}}loss_{sent}(Prob_i, K_i)+\\
    \beta_{word}*\sum_{i=1}^{Sent_{max}} \sum_{j=1}^{Word_{max}}loss_{word}(Prob_{ij}, K_{ij})
\end{multline}

Equation \ref{eq:hrnnloss} is the loss function which is the weighted sum of cross-entropy losses, $loss_{sent}$ and $loss_{word}$, where $loss_{sent}$ is the loss over probability over a sentence topic is generation $(Prob_{i})$ and $loss_{word}$ is the loss over probability over words generation $(Prob_{ij})$, with each respective sentence topic where $K$ is the paragraph description for each image $I$. The training parameters for DSIC model is such that: Sentence LSTM has $512$ units, word LSTM has $512$ units, Fully connected layer is size $1024$. Next we describe an alternative to DSIC model, TBDG model where the decoder unit takes text cues instead of image features/cues as input. 
 
\begin{figure*}[!b]
		\centering
		\includegraphics[width= \linewidth]{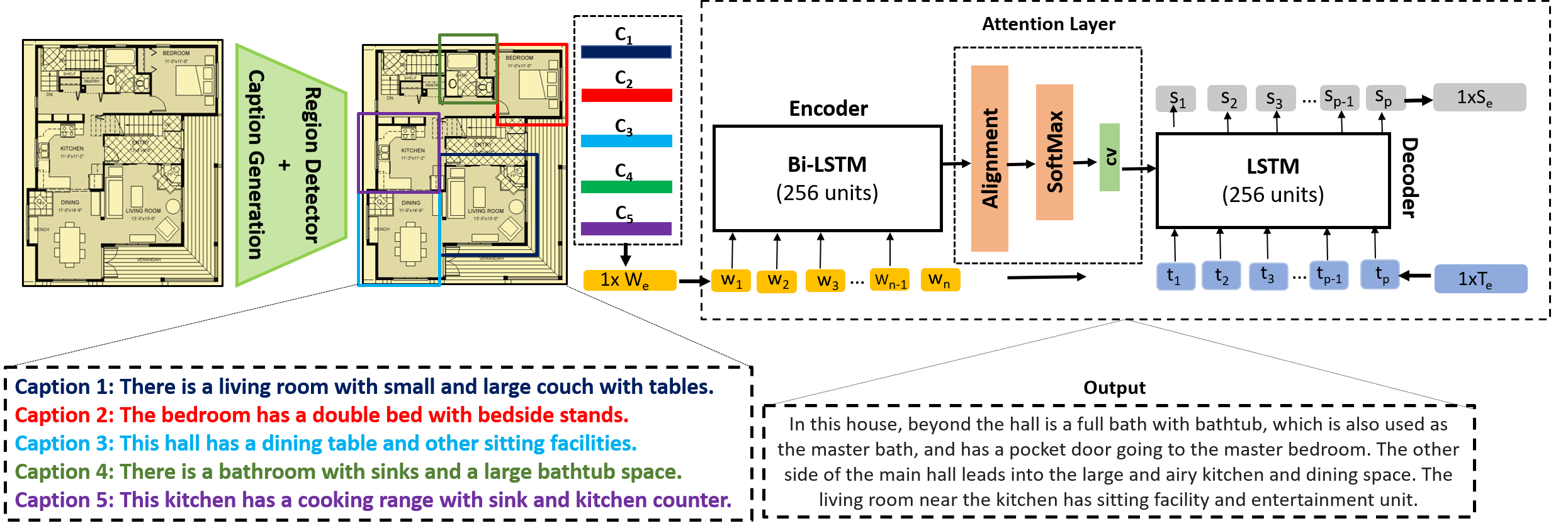}
		\caption{Framework of the proposed TBDG of generating paragraph description from input floor plan image.}
		\label{fig:framework}
\end{figure*}

\section{Transformer Based Description Generation (TBDG)}
\label{sec:TBDG}
The TBDG is a transformer based model for generating description from floor plan images. It takes input as text features by its decoder unit and generates a paragraph based description. In TBDG, RPN learns region wise captions available in BRIDGE dataset, instead of multi-sentenced paragraphs, which makes it different from DSIC model. Also, a Bi-LSTM unit acts as an encoder to the LSTM unit acting as decoder. 

\subsection{Paragraph generation with extra knowledge}
Descriptions generated directly from image cues in DSIC lack the floor plan-specific information. There are chances to miss out on salient features in the graphical document. Additional knowledge is required to generate more flexible and exciting descriptions and accurate data specific to the input image. Hence, the data available is the tuple of ($I$,$W_e$,$K$), where $I$ is the input floor plan image, $W_e$ is the word cues extracted from the image, and $K$ is the paragraph description about each floor plan. In language modeling and text generation networks, Seq2Seq models are widely used. However, with the advent of attention based seq2seq networks, popularly known as transformers, the performance of the text generation models have been increased to a great extent. In TBDG, the corpus $K$ is pre-processed for training by removing extra lines, white spaces, unknown symbols and punctuation, and tokenized using PTB tokenizer \cite{marcus1993building}. The words which are most frequently occurring are selected, and vocabulary is generated for the words. 

\subsection{Region wise captions}
Floor plan images are distinctly different from the natural images, and conventional deep models are inefficient to create features depicting a unique floor plan. Hence, learning region-wise visual features are advantageous in this context. We have extracted the region using the region proposal model described in DSIC. The annotations for regions in floor plans, available in \cite{bridge2019}, are used along with these region proposal to train an LSTM model. The model generates region-wise descriptions/captions, $C_1, C_2,...,C_n$ as shown in Fig. \ref{fig:framework}. The generated captions are taken as input to the encoder-decoder unit, which is the next stage of the pipeline, where these caption serve as extra knowledge to the decoder network. 

 \subsection{Caption fusion and word embedding generation}
At this stage, we have $n$ captions generated for each floor plan image. We select the top $5$ captions with the highest probability and fuse them as a paragraph. $C_1 \circ C_2 \circ C_3 \circ C_4 \circ C_5 = W_i$, where $W_i$ is the fused one dimensional vector of the extracted captions and $i$ is number of training sample. $W_i$ is the concatenation of word embeddings created by word2vec and $|W_e|= \min{(|W_1|, |W_2|, |W_3|,...,|W_i|)}$. Word2vec generates the embeddings for words which is a representation of each word as a vector. The dimension of concatenated vector was taken as minimum of all the vectors to avoid vanishing gradient problem during back propagation of the network.

\subsection{Paragraph encoding}
In \cite{bridge2019}, for each floor plan, a detailed paragraph description is available. However, some of the paragraphs are too long for encoding and contains additional information. Training the model with too long sequences leads vanishing gradient problem. Considering the dataset’s size, manually selecting useful information from each set of sentences is impossible. Hence, we heuristically selected a few keywords from the corpus. Examples of such keywords are common categories of regions like bedroom, bathroom, kitchen, porch, garage, and other keywords describing objects, like stairs, bathtubs, kitchen bars. From the available paragraphs, we extracted only those sentences which consist of these keywords to shorten the length of each paragraph. Each target sequence $(T_i)$ is a 1-D vector and concatenation of the word embeddings generated by word2vec, and, $|T_e|=\min{(|T_1|, |T_2|, |T_3|,...,|T_i|)} $ as shown in Fig. \ref{fig:framework}. 

\subsection{Encoder-Decoder architecture}
In TBDG model, we have proposed a transformer architecture that can handle dependencies between input and output sequence tokens by giving decoder the entire input sequence. It focuses on certain part of input sequence when it predicts the output sequence. As shown in the Fig. \ref{fig:framework}, the encoded captions $W_e$ are given as input to the Bi-LSTM unit which act as encoder. The Bi-LSTM unit generates hidden states $(h_1, h_2, h_3,...,h_n)$ and given to an attention mechanism which first generates alignment scores $e_{ij}$ between the current target hidden state $h_t$ and source hidden state $h_s$. The alignment scores are further given to SoftMax layer, which generated normalized output probabilities for each word as $\alpha_{ij}$, (See. Eq. \ref{eq:align}).
Here, $e_{ij}$ are the outputs generated by the alignment model, where $i$ is the number of time step. Attention weight $\alpha_{ij}$ is the normalized attention score at each time stamp $i$ for $j^{th}$ hidden state, where $n$ is the number of encoded words in the sentence or hidden states. Further context vector $cv_i$ is generated at every time step $i$, which is a weighted sum of encoded feature vectors. Context vector is defined as $cv_{i}= \sum_{j=1}^n(\alpha_{ij}h_i)$. Attention scores learn how relevant is the input vector to the output vector. In the Fig. \ref{fig:framework} the word embeddings, $W_e$ and $T_e$ are given as input and target output vector to the encoder unit. Equation \ref{eq:align} describes the calculation of attention scores in the proposed model. 

\begin{gather} \label{eq:align}
    e_{ij}= align(h_t, h_s)\\
    \alpha_{ij}= \frac{exp(e_{ij})}{\sum{exp(e_{in})}}  \nonumber
\end{gather}


Hence, this way decoder learns correspondence between input and output sequences in a global context and generate output sentences $S_e$. Here decoder is a LSTM network with $256$ units, connected with a Time-Distributed dense layer with SoftMax activation function. Time-Distributed dense layer applies a fully connected (dense) operation to every time step. The network parameters used for training TBDS model are such that: Optimizer used is Adam, Loss function is Categorical Cross Entropy, Input sequence length is $80$, output sequence length is kept $80$, embedding dimensions is $150$ (empirically determined).

\section{Experimental Setup}
\label{sec:exp}
Here we discuss the details of how the experiments were conducted. All the experiments  were performed on the dataset BRIDGE on a system with NVIDIA GPU Quadro $P6000$, with $24$ GB GPU memory, $256$ GB RAM. All implementations has been done in Keras with Python.

\subsection{Dataset}
\begin{figure}[t]
		\centering
		\includegraphics[scale=.23]{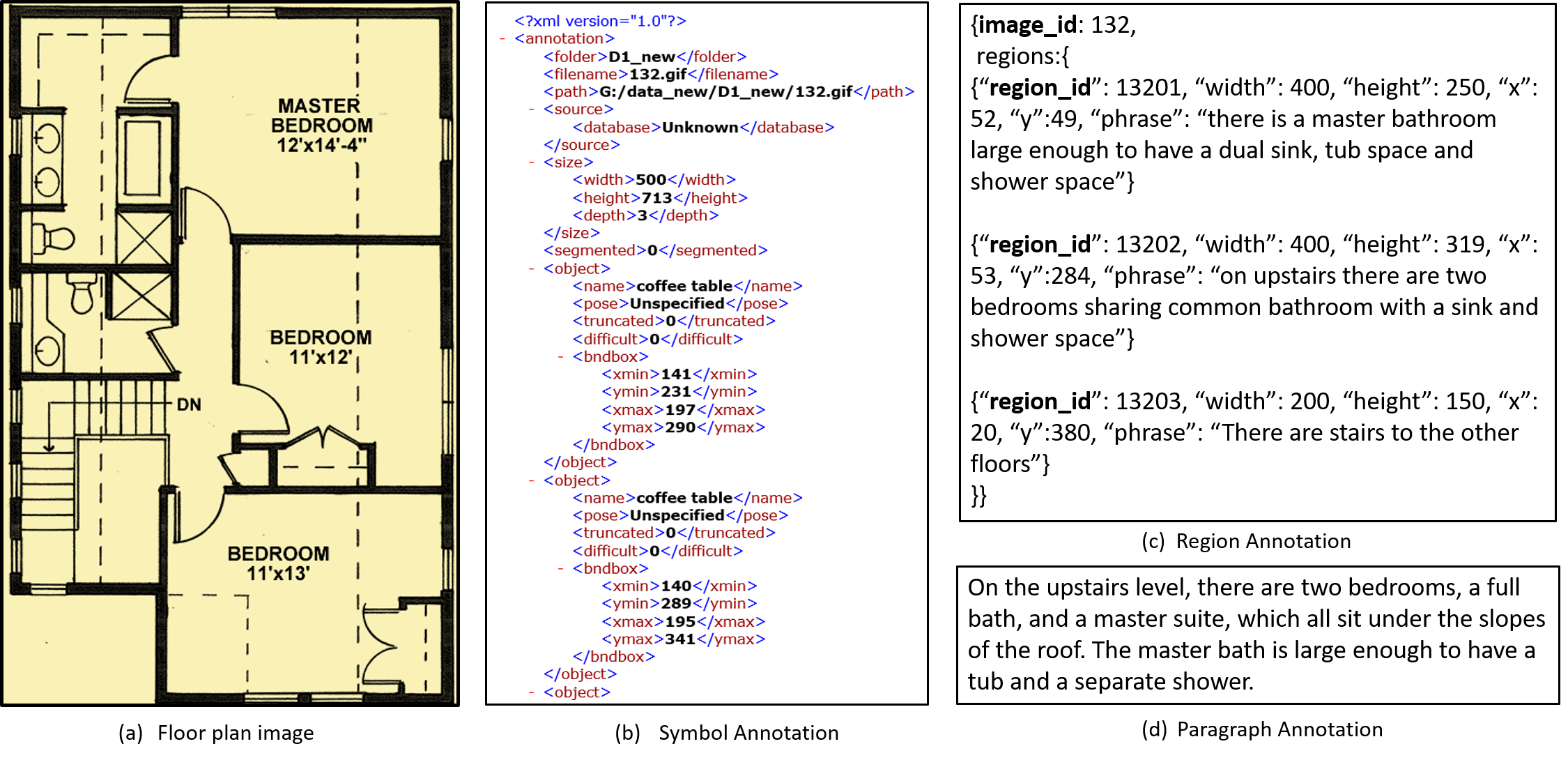}
		\caption{Image and annotations in the BRIDGE dataset \cite{bridge2019}.}
		\label{fig:dataset}
\end{figure}

In this paper, we have conducted our experiments on BRIDGE \cite{bridge2019}. This dataset has a large number of floor plan samples and their corresponding metadata. Figure. \ref{fig:dataset} shows the components of \cite{bridge2019} which has (a) floor plan image, (b) decor symbol annotations in an XML format, (c) region-wise caption annotations in JSON format, (d) paragraph based descriptions. Each paragraph’s average length in word count is $116$, with the average length of each sentence being $5$. The count of diversity is $121$, a measure of the richness of words used in sentences. There are $134942$ nouns, $5027$ verbs, $46379$ adjectives, and $5476$ proper nouns available in the dataset. 

\subsection{Quantitative Evaluation Metrics} 
\label{sec:metric}
We have quantitatively evaluated the symbol spotting accuracy and text synthesis quality. The performance metrics are defined next.

1. ROUGE:  It is a set of metrics designed to evaluate the text summaries with a collection of reference summaries. We have compared the generated descriptions with available human-written descriptions using n-gram ROUGE based on the formula 
\begin{equation}
    \frac{\sum_{S\in\{RS\}}{\sum_{gram-n\in S}{Count_{m}(gram-n)}}}{\sum_{S\in\{RS\}}{\sum_{gram-n\in S}{Count(gram-n)}}}
\end{equation}
where $RS$ stands for reference summaries, $n$ stands for length of the n-gram, $gram-n$, and $Count_{m}(gram-n)$ is the maximum number of n-grams co-occurring in the candidate summary and the set of reference summaries.

2. BLEU:
It analyses the co-occurrences of n-grams between a machine translation and a human-written sentence. The more the matches, the better is the candidate translation is. The score ranges from $0$ to $1$, where $0$ is the worst score, and $1$ is the perfect match.  The n-gram modified precision score $(p_n)$ is computed as:
\begin{equation}
 p_n=\frac{\sum_{C\in \{Cand\}}{\sum_{gram-n\in C}{Count_{clip}(gram-n)}}}{\sum_{C'\in\{Cand\}}{\sum_{gram-n'\in C'}{Count(gram-n')}}} \nonumber
\end{equation}
$Count_{clip}$ limits the number of times an n-gram to be considered in a candidate ($Cand$) string. Then they computer the geometric mean of the modified precision $(p_n)$ using n-gram up to length $N$ and weight $W_n$, which sums up to $1$. A brevity penalty (BP) is used for longer candidate summaries and for spurious words in it, which is defined by the following equation:
 \begin{equation}
    BP=
    \begin{cases}
      1, & \text{if}\ c>r \\
      e^{\frac{1-r}{c}}, & c\leq r
    \end{cases}
  \end{equation}
$c$ is the length of the candidate summary, and $r$ is the length of the reference summary. Then BLEU score for corpus level given equal weights to all n-grams is evaluated by the following equation:
\begin{equation}
    BLEU=BP.exp^{\sum_{i=1}^{N}{W_n}log(p_n) }
\end{equation}
Here $W_n$ is the equally distributed weight in n-grams. E.g. in case of BLEU-$4$ the weights used are $\{(0.25),\\(0.25),(0.25),(0.25)\}$.

3. METEOR: It is a metric used for evaluating machine-generated summaries with human-written summaries by checking the goodness of order of words in both. METEOR score is a combination of precision, recall, and fragmentation (alignment) in the sentences. It is a harmonic mean of the uni-gram precision and uni-gram recall given alignment and calculated as:
\begin{gather}
    PN=\frac{1}{2}\left(\frac{no \;of\; chunks}{matched\; uni-grams}\right)\\
    METEOR=\frac{10PR}{R+9P}(1-PN)
\end{gather}
$PN$ is the penalty imposed based on a larger number of chunks, $P$ are the uni-gram precision, $R$ is the uni-gram recall. METEOR is the final score obtained by multiplying the harmonic mean of unigram precision and uni-gram recall with the penalty imposed.

4. Average Precision (AP): The metric average precision used for evaluating the performance of decor symbol detection method is defined by the following equation:
\begin{equation}
    AP= \frac{1}{N_s}* \sum P_r(rec)
\end{equation}
Where, $N_s$ is the total detection for each class of symbol, $P_r$ is the precision value as a function of recall$(rec)$. Mean average precision (mAP) is the average of all the $AP$ calculated over all the classes. 

\begin{figure}[t]
		\centering
		\includegraphics[scale=.30]{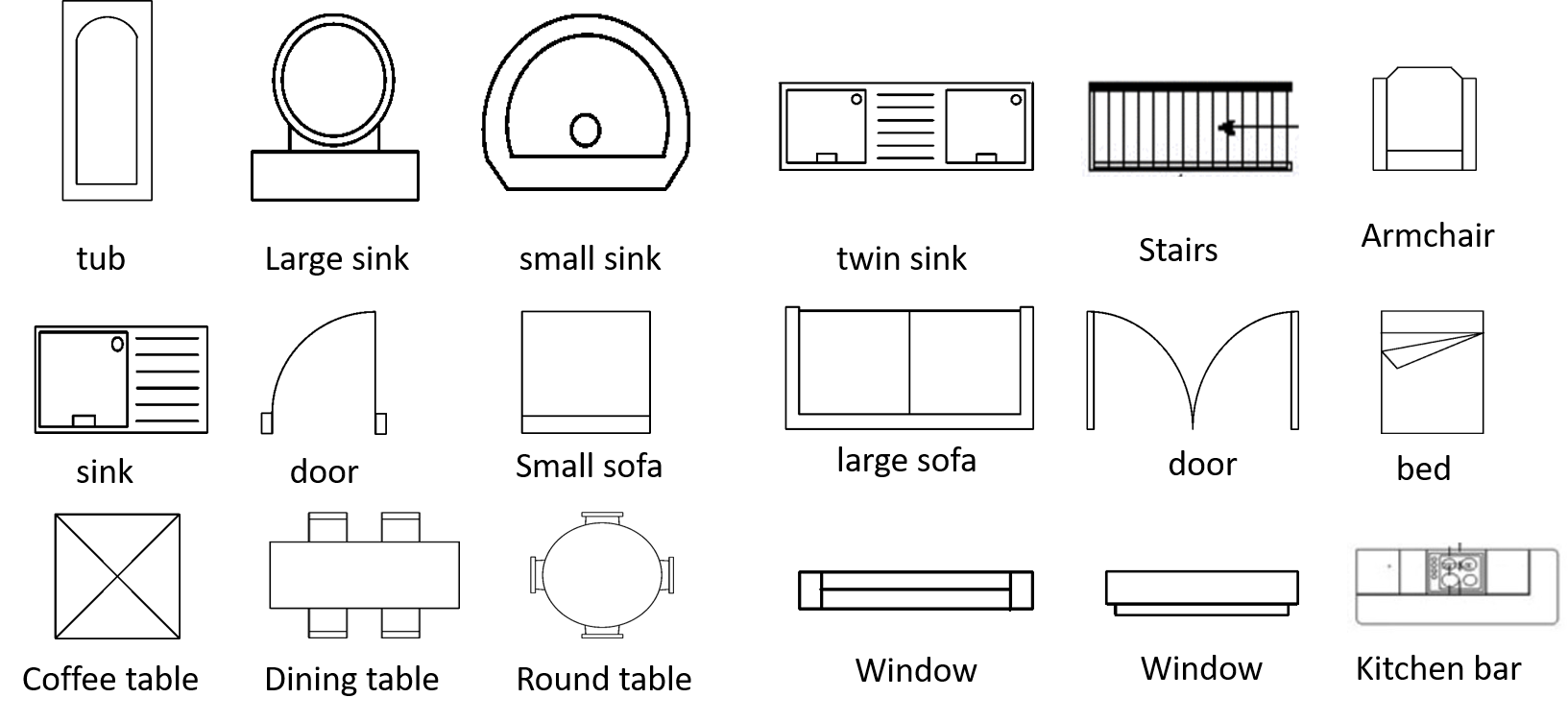}
 		\caption{The object classes available in BRIDGE dataset \cite{bridge2019}.}
		\label{fig:obj}
\end{figure}
\section{Results of the proposed models}
\label{sec:baseline}
In the next sections, results generated with the proposed models are described in detail. To validate the superiority of the proposed models DSIC and TBDG, the description generation by a multi-staged pipeline with deep learning is also proposed and a comparative analysis is done to validate the choice of the networks used. In the next sections, steps involved in visual element detection are described in detail. It also discusses the resultant detection and classification of visual elements in the proposed pipeline. 
\subsection{ Decor symbol detection and classification}
\begin{figure}[t]
		\centering
		\includegraphics[scale=.3]{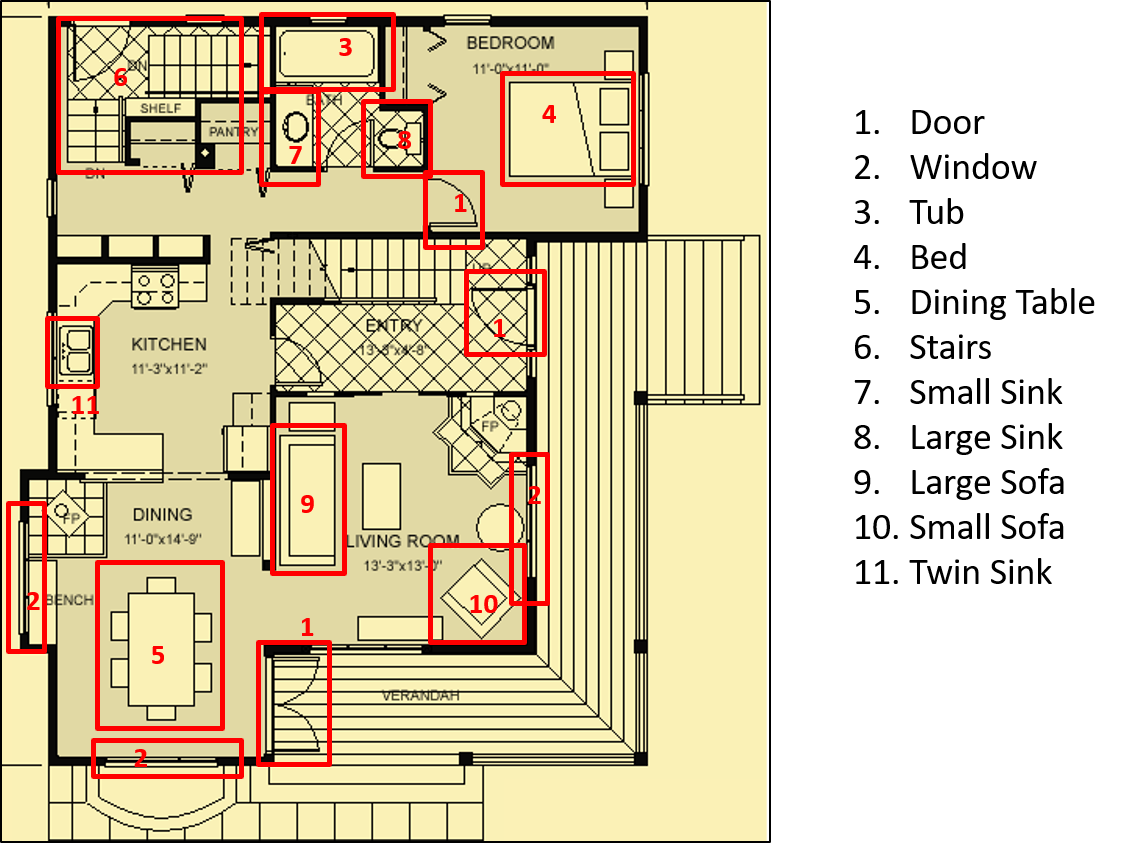}
		\caption{The spotted decor symbols in a given floor plan.}
		\label{fig:Sym}
\end{figure}

Symbol spotting is a widespread problem in document image interpretation. In \cite{bridge2019}, there are annotations for the decor symbols. In this work, we have adapted the YOLO model \cite{redmon2017yolo9000} for detecting and classifying the decors by fine-tuning it using the decor symbol annotations present in the BRIDGE dataset. The symbol spotting network has $9$ convolutional layers with max pool layers in between and is fine-tuned for $16$ object categories (as shown in Fig. \ref{fig:obj}). The trained network has $105$ filters (for BRIDGE dataset) and a linear activation function.
The predicted class confidence score is calculated as $Prob(object) \times IoU$ . Here, $IoU$ is the intersection of union between the predicted bounding box and the ground truth bounding box. It is calculated as $IoU= \frac{Area\ of\ Intersection}{Area\ of\ Union}$. 
 At the same time, $Prob(object)$ is the probability of detecting the object in that bounding box.  The decor symbol detected here $(o_i)$, are used in generating semi-structured description in later stage. The decor symbols in floor plans can vary widely because of the representation across different datasets. Also, in the real world floor plans made by architects, the model might differ. We introduced variability by including samples of floor plans from different datasets such as \cite{delalandre2010generation,de2015cvc,sharma2017daniel} for decor symbol annotations. The training dataset covers a wide range of decor symbols, making the network detect and recognize the symbols’ variability. The detected decor symbols in floor plan images are shown in Fig. \ref{fig:Sym}. The two images are taken from BRIDGE datasets and shows variability in decor symbol for two different floor plan images. A wide variability in decor symbols in included in training dataset in order to make the detection model more general. The symbols which are not detected for example, ``billiard" and ``cooking range", are not included in symbol annotations.  

\begin{figure}[H]
		\centering
		\includegraphics[width= \linewidth]{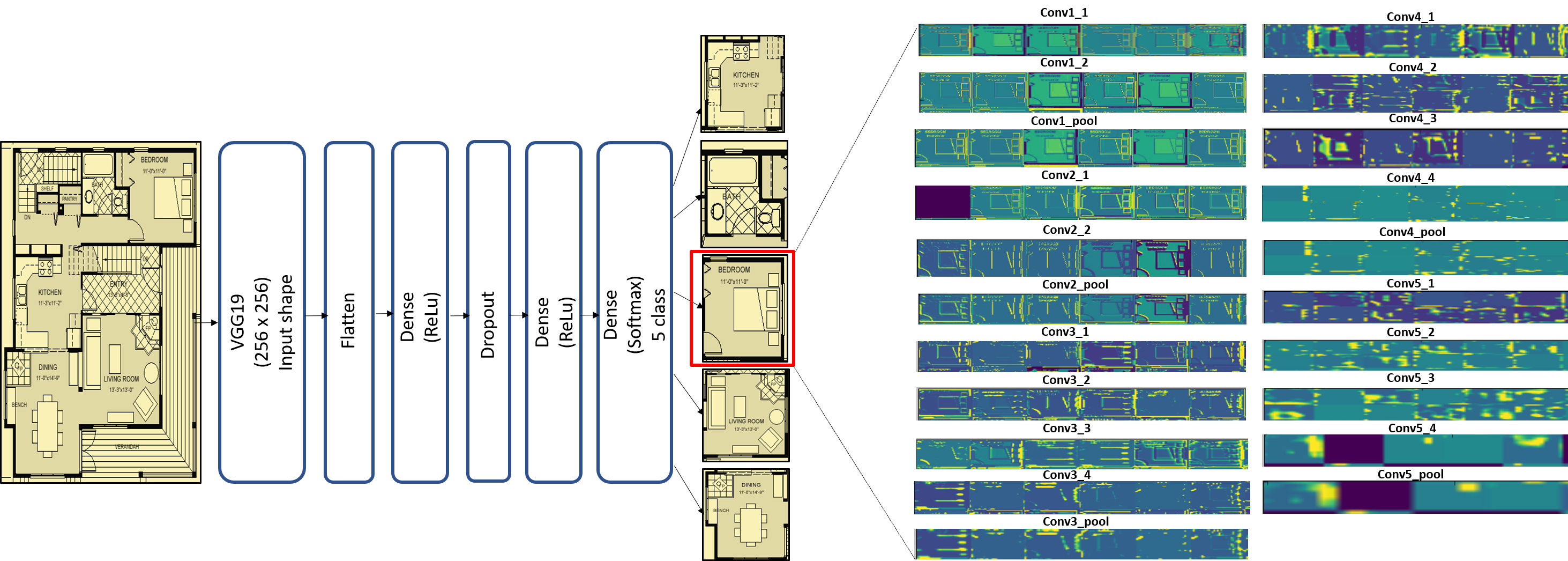}
		\caption{Image classification network with top-6 activation maps from each layers.}
		\label{fig:imclass}
\end{figure}
\begin{figure}[H]
		\centering
		\includegraphics[width= \linewidth]{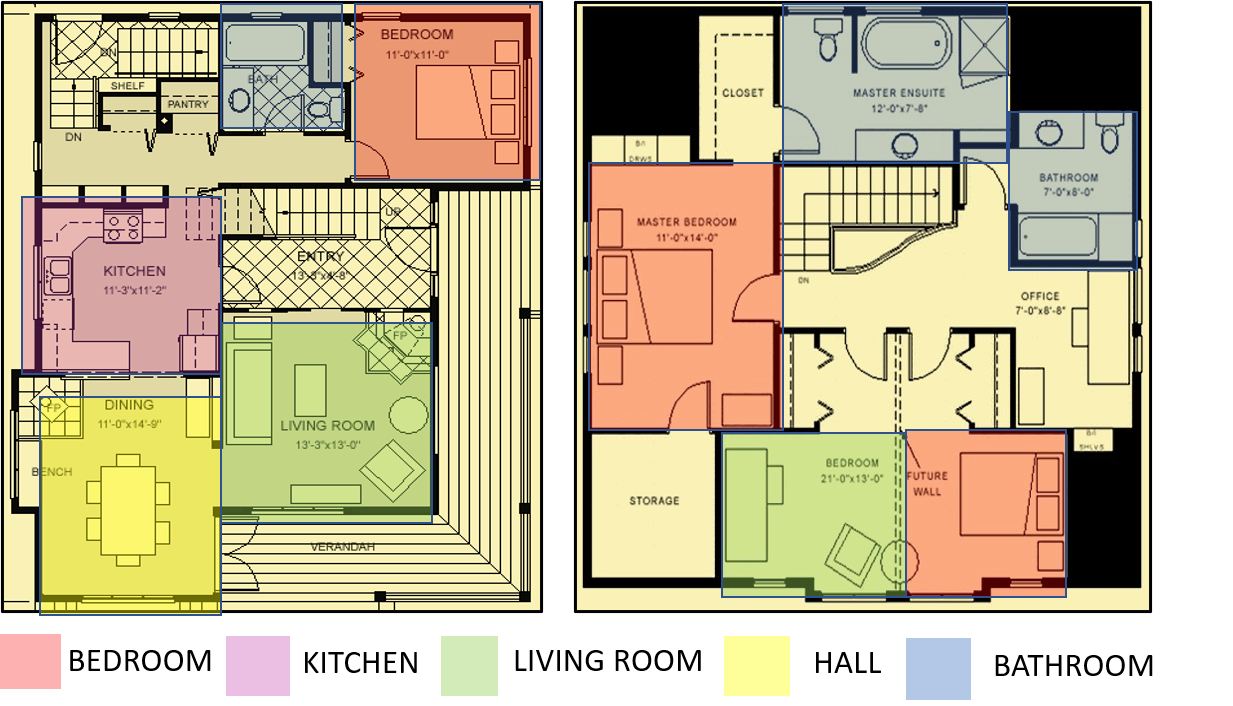}
		\caption{{Room classification results for $5$ classes.}}
		\label{fig:room_class}
\end{figure}
\subsection{Room characterization} 
\label{sec:imclass}
As a step of multi-staged floor plan recognition, room characterization is a step to recognize and classify individual room in a floor plan to its respective class. In this regard, rooms in each floor plan are classified in $5$ classes, \textit{Bedroom, Bathroom, Kitchen, Hall, Living room}. Annotations for each room class are taken from BRIDGE dataset, where region bounding boxes are available and class names are taken from the region wise captions for respective bounding box. 
A deep learning image classification model using VGG19 as a backbone network is used as a classification framework. Figure \ref{fig:imclass} shows the framework diagram of the model used and image of each room class along with activations for a sample class image, Bedroom. In this network, only the last $5$ layers are kept trainable in VGG19 and appended with two dense and dropout $(0.5)$ layers. 
Figure \ref{fig:imclass} depicts that activations for initial layers retain almost all the information from the image, focusing on specific parts such as edges and the image’s background. However, in the deeper layers, activations are less visually interpretable. All the characterized rooms$(r)$ from an input floor plan image are stored as $(r_1,r_2,...,r_n)$. Figure. \ref{fig:room_class} shows the resultant classification for floor plan images room classification framework into $5$ defined classes. The different colors for different classes are shown in the legend. The empty spaces in the floor plan are not marked as any room class in the BRIDGE dataset, hence they are not classified by the model.
VGG19 pre-trained on ImageNet dataset is fine-tuned with a $1920$ training sample of $5$ room classes, and validation is done over $460$ samples. The training data contains a mixed sample of room images from \cite{bridge2019,de2015cvc,sharma2017daniel}. The rooms $r_i$, generated here are used in generating multi-staged description in the later stage. The number of samples for each class in training and validation dataset are: \textit{Bedroom:} $440$ and $86$, \textit{Bathroom:} $887$ and $223$, \textit{Kitchen:} $287$ and $72$, \textit{Hall:} $75$ and $21$, \textit{Living Room:} $231$ and $58$ in respective order.


\subsection{Description generation }
\label{sec:desgen}
\begin{figure}[H]
		\centering
		\includegraphics[scale= 0.4]{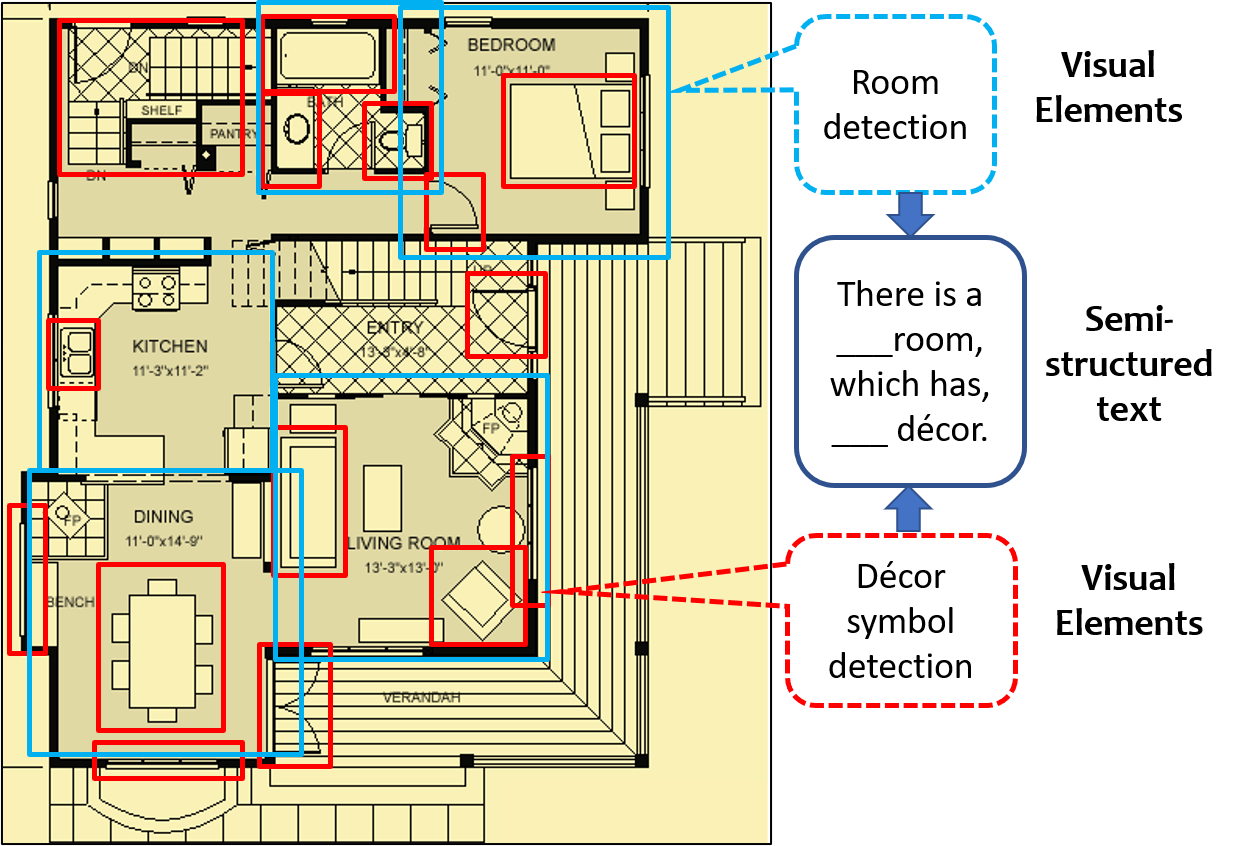}
		\caption{Semi-structured description generation.}
		\label{fig:template}
\end{figure}
\begin{figure}[!b]
		\centering
		\includegraphics[width= \linewidth]{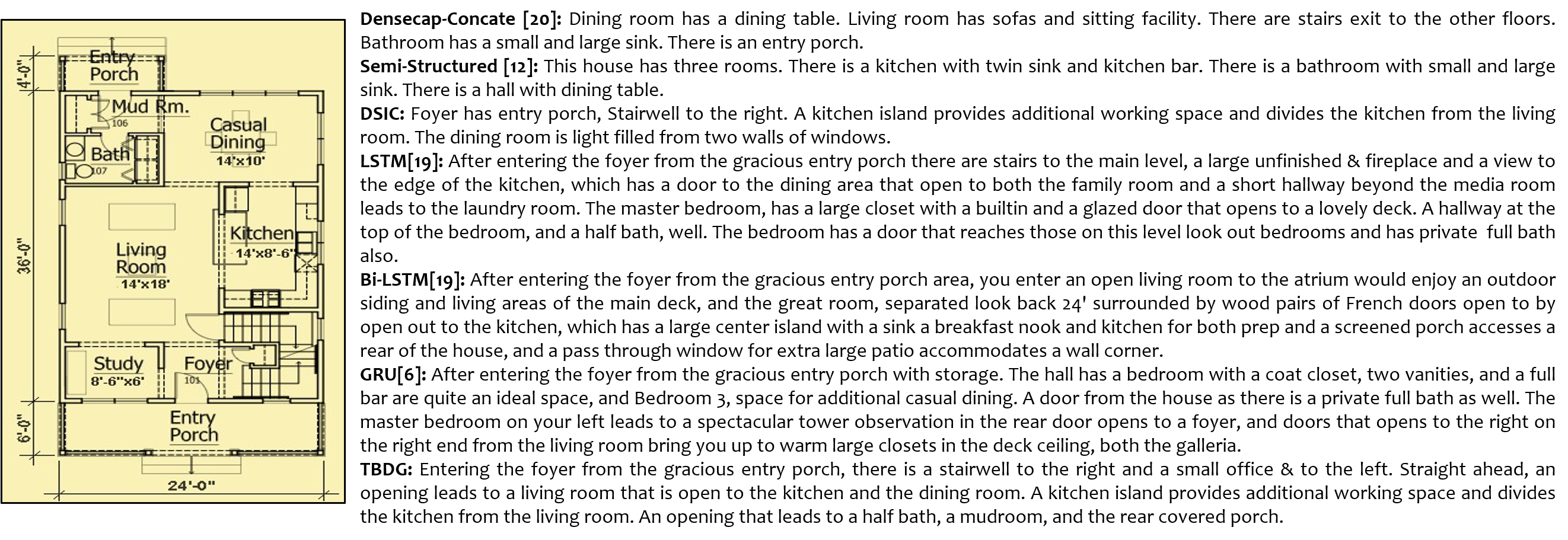}
		\caption{Descriptions generated with proposed models and various baseline models.}
		\label{fig:result_1}
\end{figure}
In the previous sections, different visual elements from the floor plans are detected and classified using various deep learning models. In the multi-staged pipeline, these visual elements are used with a semi-structured sentence model proposed in \cite{goyal2018asysst}, \cite{goyal2019sugaman}, and a description for the given floor plan is generated. Figure \ref{fig:template} depicts an example where the synthesized descriptions for a given floor plan image with the visual elements described in the previous steps. Figure \ref{fig:result_1} shows the results generated with the proposed models TBDG, DSIC, semi-structured sentence based model, and other baseline models. 
In this paper, a comparison of semi-structured sentences with the learned sentences is presented with depicts the superiority of end-to-end learning models with the multi-staged pipelines. Multi-staged pipeline for floor plan recognition and description generation is presented here as a comparison with the end to end models DSIC and TBDG. Multi-staged pipelines have been used in literature for floor plan recognition and description generation in \cite{goyal2018asysst,goyal2019sugaman,goyal2018plan2text,madugalla2020creating}. In multi-stage pipelines, the accuracy of the generated descriptions depends upon the accuracy of the intermediate stages. Hence, miss-classification of one component will lead to error in the output sentence. This rationale is the driving factor to come up with an end-to end learning model with advanced deep neural networks. In the next sections, comparative analysis for various modules and sub-modules are discussed in details, along with qualitative and quantitative evaluation of generated descriptions. 



\section{Comparative analysis with state of the art}
\label{sec:des_gen}
In this section, comparative analysis with various state of the art description generation schemes and sub-stages involved are presented with their qualitative and quantitative evaluation. It describes the performance evaluation of proposed models and several baselines with metrics discussed in Sec. \ref{sec:metric}. 

\subsection{Comparative analysis in multi-staged pipeline} 
\label{sec:visuals}
In this sub-section, we present how the various stages of multi-staged pipeline performed as the performance evolution of various stages. We also performed a quantitative comparison of various steps involved in multi-staged pipeline in order to validate the choice of network used.

\subsubsection{Decor Identification:}
\begin{figure}[H]
		\centering
		\includegraphics[width=\linewidth]{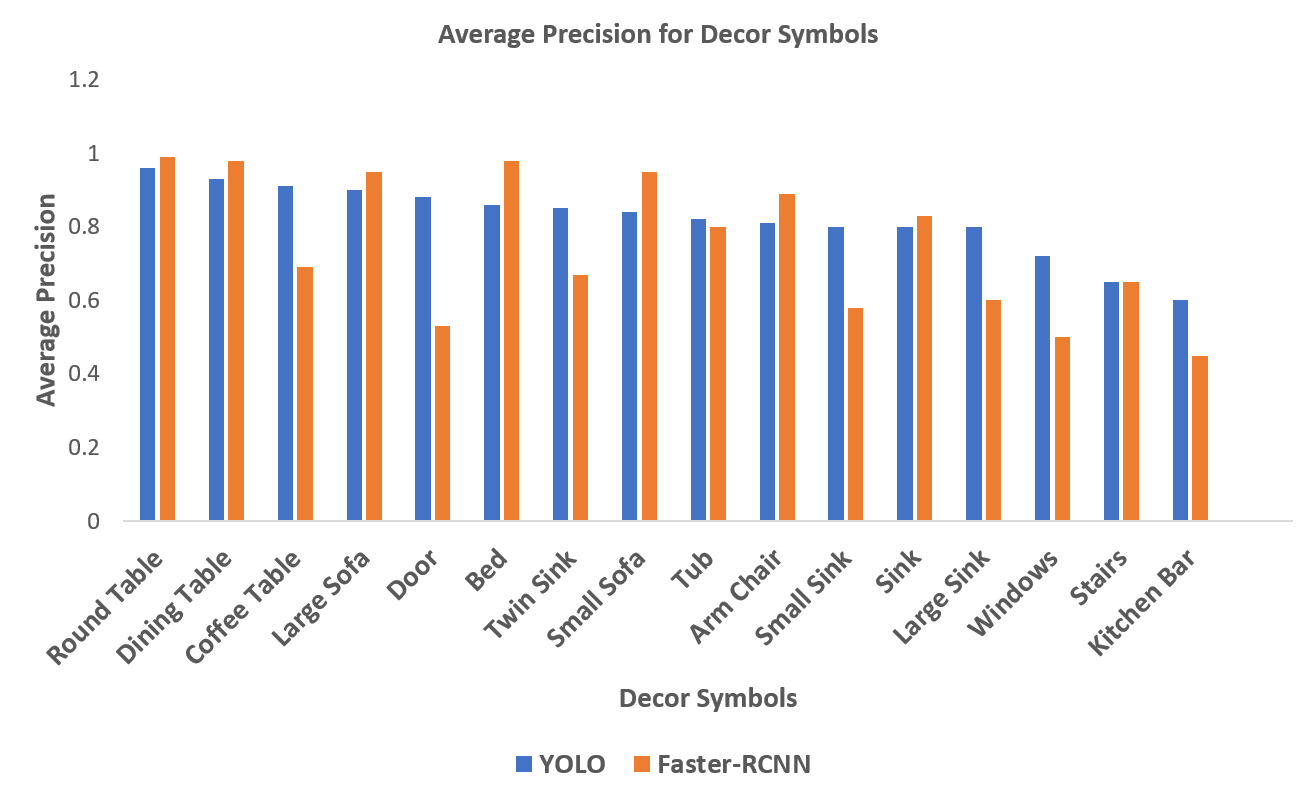}
		\caption{Decor identification with YOLO and Faster-RCNN.}
		\label{fig:symb_detect}
\end{figure}
Figure \ref{fig:symb_detect} shows a comparative analysis of YOLO and F-RCNN trained on BRIDGE dataset. The mAP obtained for decor symbol spotting network using YOLO is $82.06\%$  and for F-RCNN is $75.25 \%$. For a few categories of symbols, F-RCNN is performing better, but overall mAP is $\sim 7\%$ higher for YOLO. Hence, YOLO, is used in model instead of Faster-RCNN given the better performance. Also in the work \cite{rezvanifar2020symbol}, symbol spotting from architectural images is done for occluded and cluttered plans using YOLO, concluding the fact that YOLO as a single shot detector performs better than two stage classification networks such as Faster-RCNN for architectural drawings. 

\begin{figure}[H]
		\centering
		\includegraphics[width= \linewidth]{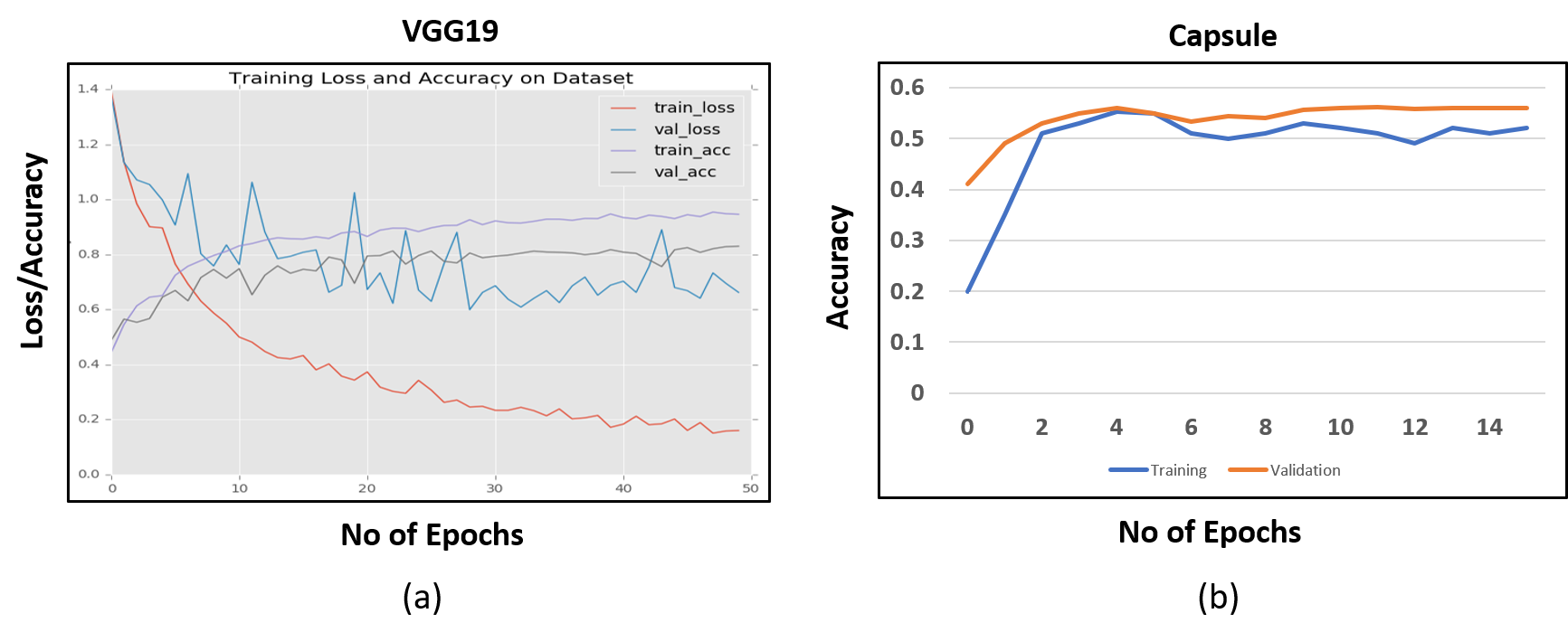}
		\caption{Performance evaluation of room classification.}
		\label{fig:curves1}
\end{figure}

\subsubsection{Room Characterization:}

Figure. \ref{fig:curves1} (a),(b) depicts the performance of image cues/ visual elements extraction from floor plan images for room classification. Figure. \ref{fig:curves1}(a) is the training and validation accuracy and loss curves for room image classification using VGG19 backbone network. 
After training for $50$ epochs, an accuracy of $82.98\%$ could be achieved in-room image classification. The fluctuation of validation loss is due to the uneven distribution of the number of images in all $5$ classes. A $5$-fold cross-validation over the data samples was performed on training data to validate the model.  

The room image classification model discussed in Sec. \ref{sec:imclass}, was also implemented using much recent Capsule network \cite{sabour2017dynamic} as a backbone network, which gave a classification accuracy of $56.01 \%$, making VGG19 the obvious choice for the backbone network. Figure. \ref{fig:curves1}(b) is the training and validation accuracy for room classification model using Capsule network.
The performance of the room characterization on BRIDGE dataset was also tested with classical machine learning methods proposed in \cite{goyal2019sugaman,goyal2018asysst}. BoD classifier with multi-layered perceptron, proposed in \cite{goyal2018asysst}, gave a validation accuracy of $61.30 \%$ and LOFD proposed in \cite{goyal2019sugaman} with multi-layered perceptron gave $63.75 \%$ of accuracy, while the validation accuracy of proposed model is $82.98\%$, making VGG19 a suitable choice for room classification model. In Fig. \ref{fig:room_class}, the two images have variability in the representation of each room class, but features learnt using convolutional networks are much robust in case of variable representation of images of same class as compared to hand-crafted features, leading to higher validation accuracy.

\subsection{Quantitative evaluation of description generation}
In this sub-section we present the quantitative evaluation of our proposed model with other state-of-the-art. The baseline descriptions are generated using language modelling where models such as LSTM \cite{hochreiter1997long}, Bi-LSTM \cite{hochreiter1997long}  and GRU \cite{cho2014properties} are experimented with. Language modeling is done by learning an entire corpus. These paragraph corpus are the textual descriptions for floor plans \cite{bridge2019}. The generated descriptions from the proposed models and presented baselines are compared on various matrices defined in Sec. \ref{sec:metric} and the quantitative results are presented in the Tab. \ref{tab:score}. 

\begin{figure}[H]
		\centering
		\includegraphics[width= \linewidth]{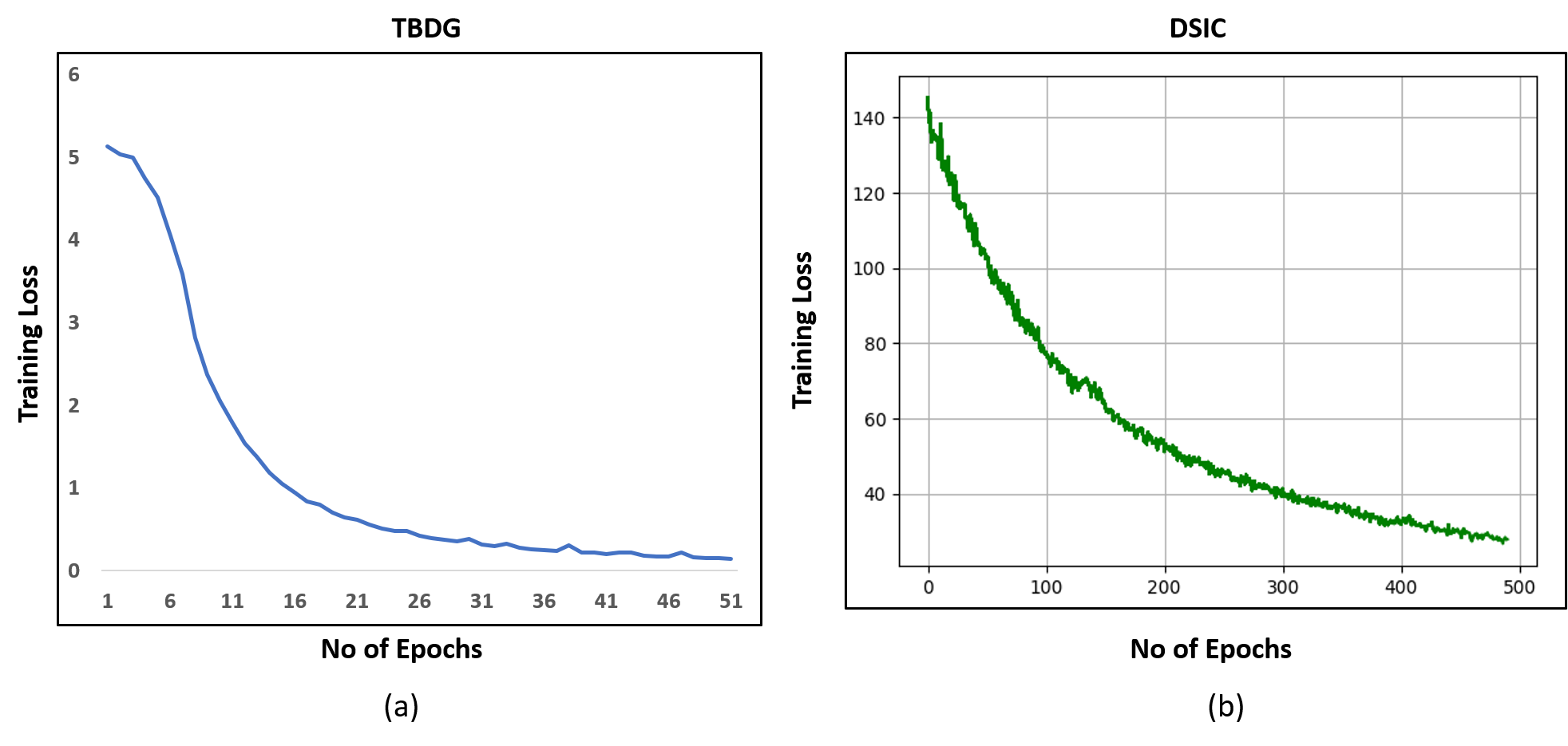}
		\caption{Performance evaluation for TBDG \& DSIC models.}
		\label{fig:tbdg_eval}
\end{figure}

\begin{figure}[H]
		\centering
		\includegraphics[width=\linewidth]{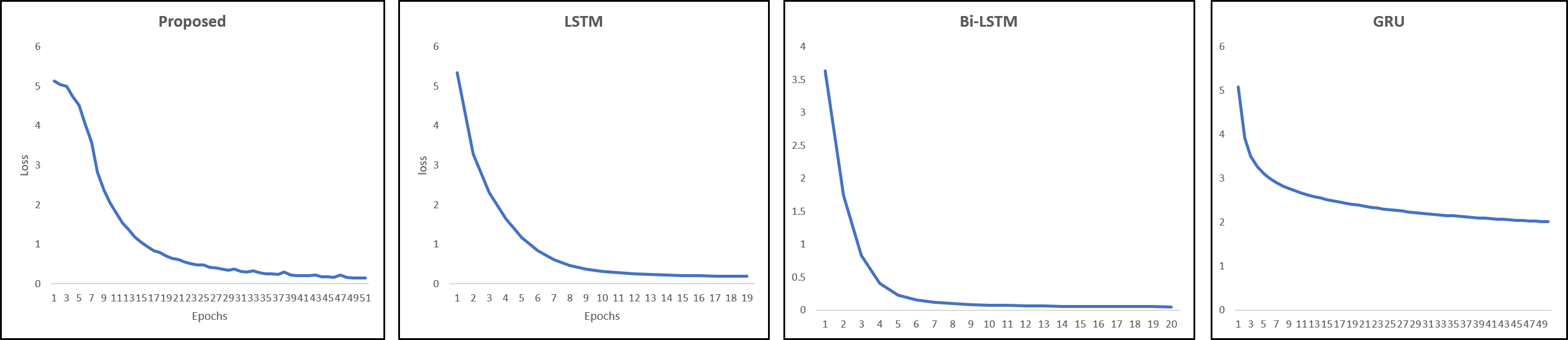}
		\caption{Loss curves of different language models.}
		\label{fig:curves2}
\end{figure}

 Figure. \ref{fig:tbdg_eval}(a) shows the loss curve for TBDG for the part of language learning (Sequence2Sequence training, LSTM as encoder, Bi-LSTM as decoder), Figure. \ref{fig:tbdg_eval}(b) shows the loss curve for DSIC for language learning part (CNN as encoder, hierarchical RNN as decoder). In TBDG, since LSTM and Bi-LSTM layers are used for training with word features, in the network, the loss converged below $1$ and became stable in $51$ epochs. In DSIC, training LSTM based, hierarchical RNN with image features took longer epoch time to converge than TBDG because of larger number of trainable parameters. 
Figure  \ref{fig:curves2}(a), (b), (c) shows loss curves for all the baseline language models for language modelling, i.e. LSTM, Bi-LSTM and GRU models respectively. As it can be seen that the loss value reached below $1$ but did not became $0$ while training for $20$ epochs. Also GRU has the similar benefits but they are more efficient than LSTMs when training with more data is required. In this case, LSTM and Bi-LSTM took $\sim 550$ ms/epoch, while GRU took $\sim 300$ ms/epoch. The loss value in GRU got stabilize earlier than LSTMs while trained for $50$ epochs.

\begin{table*}[t]
\centering
\caption{Evaluation of generated paragraphs with different metrices (METEOR, BLEU, ROUGE).}
\resizebox{\columnwidth}{!}{
\begin{tabular}{ ||c| c| c| c| c| c|c |c|c|| }
\hline
 \multirow{2}{*}{\textbf{Method}} & \multirow{2}{*}{\textbf{BLEU-1}} & \multirow{2}{*}{\textbf{BLEU-2}} & \multirow{2}{*}{\textbf{BLEU-3}} & \multirow{2}{*}{\textbf{BLEU-4}} & \multirow{2}{*}{\textbf{METEOR}}& \multicolumn{3}{c||}{$ROUGE_L$}\\
 \cline{7-9}
 &&&&&&\multicolumn{1}{c|}{\textit{precision}}&\multicolumn{1}{c|}{\textit{recall}}&\multicolumn{1}{c||}{\textit{f-score}}
 \\  \hline
 \textbf{Densecap-concat} &$0.1353$ & $0.0586$ & $0.0955$ & $0.2373$ & $0.0530$ &$0.9416$ &$0.3322$ &$0.4910$\\  \hline
 \textbf{Semi-Structured} &  $0.1519$ & $0.1613$ & $0.1622$ & $0.432$ & $0.0677$& $0.9215$ & $0.3410$ & $0.4977$ \\ \hline
 \textbf{DSIC} &  $0.7013$ & $0.6794$ & $\mathbf{0.6637}$ & $\mathbf{0.6543}$ & ${{0.4460}}$ &$1.4797$ & $1.0593$ & ${1.2346}$ \\ \hline
 \textbf{LSTM} &  $0.4464$ & $0.3048$ & $0.2166$ & $0.1673$ & $0.2076$ &$0.7648$ & $0.6063$ & $0.6763$ \\ \hline
 \textbf{Bi-LSTM} &  $0.4629$ & $0.3058$ & $0.2275$ & $0.1699$ & $0.2281$& $0.6852$& $0.6880$ & $0.6865$   \\ \hline
 \textbf{GRU} &  $0.4487$ & $0.3019$ & $0.2194$ & $0.1691$ & $0.1825$& $0.6261$ & $0.6892$ & $0.6561$ \\ \hline
 \textbf{TBDG} &  {$\mathbf{0.7277}$} & $\mathbf{0.6866}$ & $0.6633$ & $0.6326$ & $\mathbf{0.4927}$& $\mathbf{1.5283}$ & $\mathbf{1.1142}$ & ${\mathbf{1.2867}}$ \\ \hline
\end{tabular}
}
\label{tab:score}
\end{table*}

\begin{figure}[H]
		\centering
		\includegraphics[width= \linewidth]{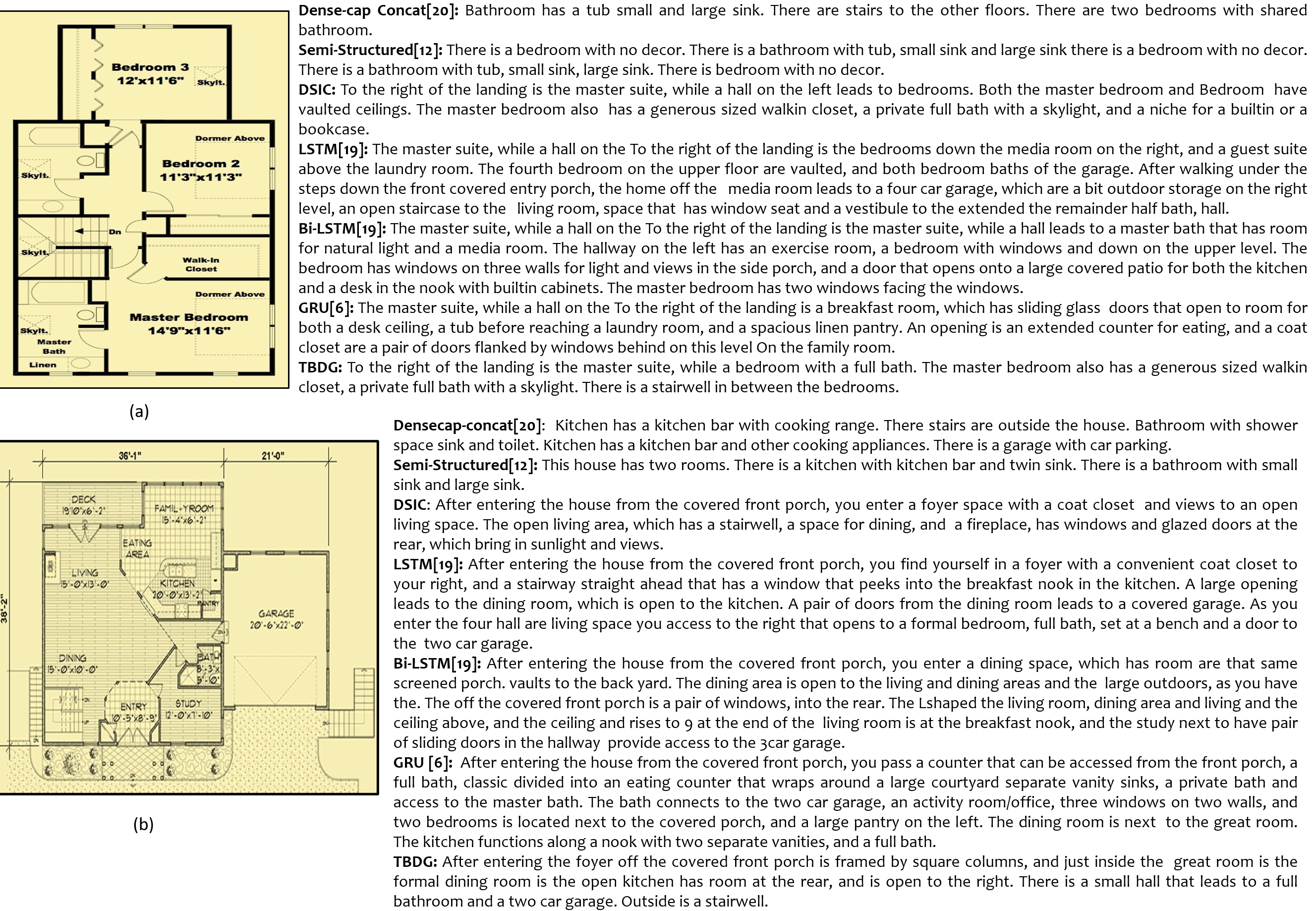}
		\caption{Descriptions generated with proposed models and various baseline models.}
		\label{fig:result_im1}
\end{figure}

Table \ref{tab:score} shows the quantitative comparison of description synthesis with the proposed models and the presented baseline models for various metrics with the ground truth paragraphs available in \cite{bridge2019} where the values in bold, represents the highest value of a particular metric for a given model. The evaluation is done on BLEU-\{$1$,$2$,$3$,$4$\}, $ ROUGE_L $, and METEOR, where the BLEU score variant depends upon the n-gram. It can be seen that the performance of TBDG is better than all other description generation schemes on all the metrics except for BLEU-$3,4$. It is least in Semi-Structured template-based method, and Densecap-concatenated paragraphs (taking top $5$ sentences from Densecap model trained on floor plans) since the sentences have a fixed structure given different input images. However, performance increase for the language models, LSTM, Bi-LSTM, and GRU even if they do not generate image specific sentences. These language models generate phrases and context used in the training corpus while generating sentences when we use a seed sentence to create a paragraph, which increases the BLEU scores for different n-grams. $ROUGE_L$ also gives the highest precision-recall and f-score values for the TBDG model. Hence, it can be concluded that the knowledge-driven description generation (TBDG) performs better than generating descriptions directly from image cues (visual features). Other language models generate sentences using corpus phrases but not specific to the input image, which is not very useful in the current scenario. The qualitative evaluation and comparison of the the proposed models with the baseline models are discussed next.

\begin{figure}[H]
		\centering
		\includegraphics[width= \linewidth]{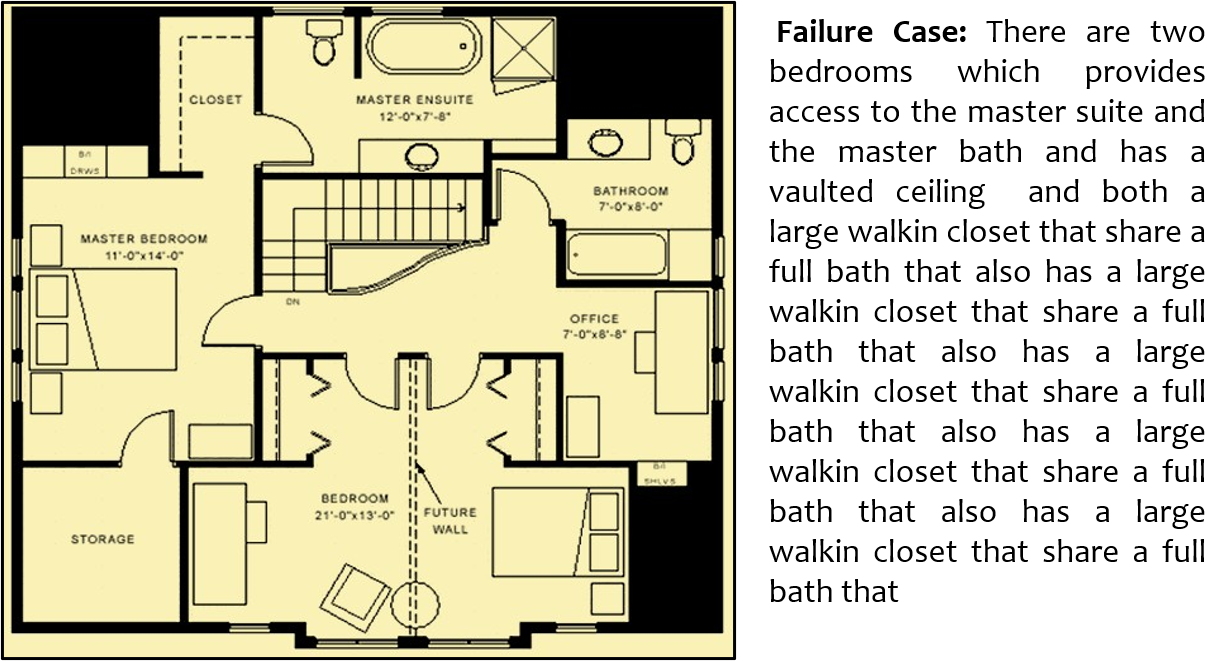}
		\caption{Failure case with the TBDG model.}
		\label{fig:result_im2}
\end{figure}

\begin{figure}[H]
		\centering
		\includegraphics[width= \linewidth]{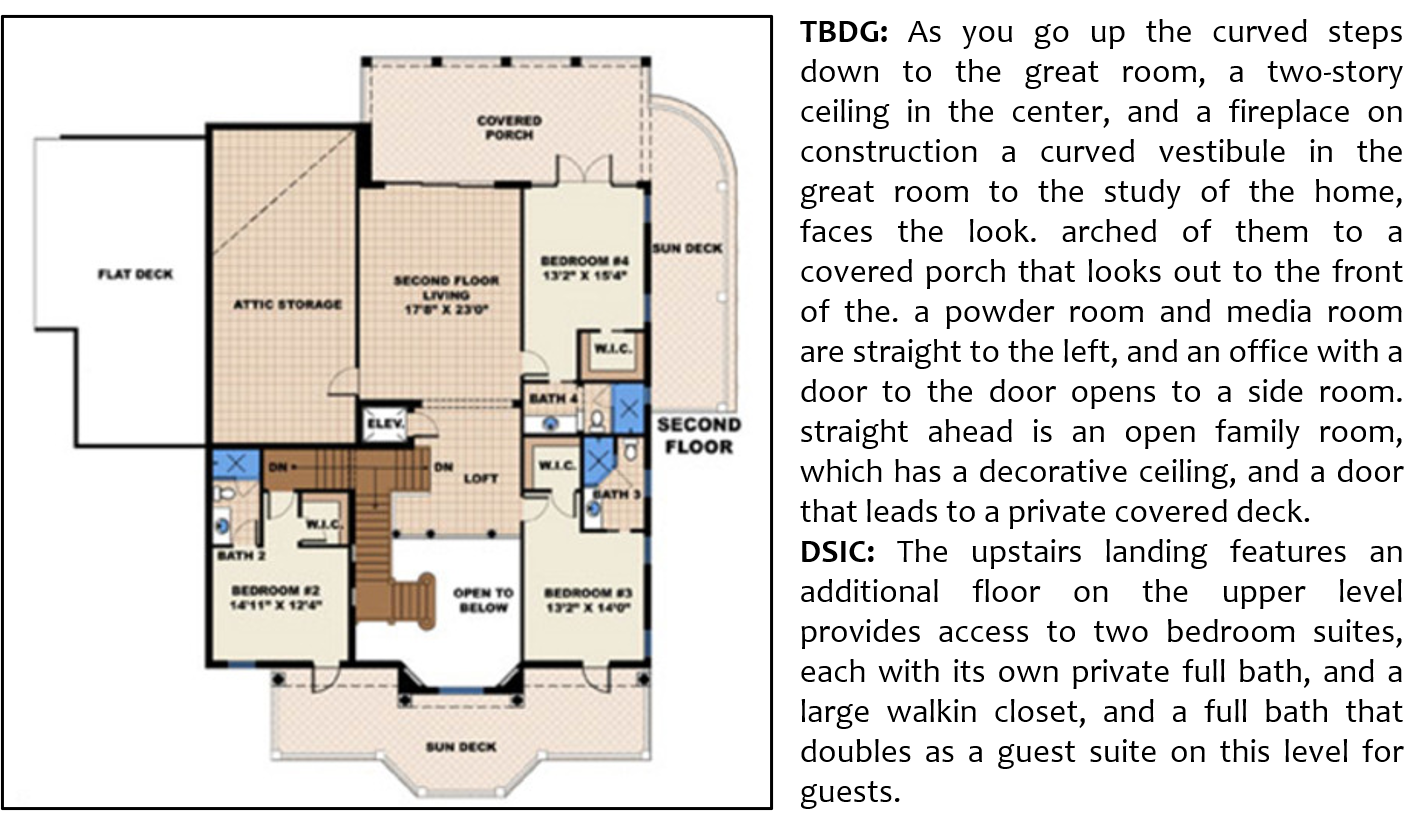}
		\caption{An illustration depicting the robustness of the TBDG model over DSIC for a general floor plan image.}
		\label{fig:result_im3}
\end{figure}
\subsection{Qualitative evaluation of description generation}
All the paragraph descriptions generated by various techniques are shown in Fig. \ref{fig:result_im1}, which are corresponding to the images shown with respective set of descriptions.
Results show that paragraphs generated by \cite{goyal2019sugaman} and \cite{johnson2016densecap}  are simple and have a fixed structure and they do not have flexibility. They do not describe connection of a room with another in a global context.
However, paragraphs generated from DSIC and TBDG are very descriptive and close to human written sentences. They also include specific details of the images, for example details about contents of a bedroom, such as closets and bathrooms, details about staircase in a hall. They also include details about other areas in a floor plan for example porch and garage, which multi-staged based methods fails to describe because they do not have these room classes included in their training data. These models themselves capture intricate details in the descriptions, in which multi-staged based methods fail, since they require explicit annotation for every component. 
Also, paragraphs generated from other baseline, LSTM, Bi-LSTM, GRU language models, are generating phrases and words related to the language structure but possess very less relevance to the input image. Hence, these kinds of models are suitable for poetry, story, and abstract generation but not for an image to paragraph generation. 

Figure. \ref{fig:result_im2} shows the failed prediction of paragraph for the proposed model TBDG. 
Sometimes the model fails to generate longer sequences or the words which are less frequent in the vocabulary, and then it starts repeating the sentences. Figure. \ref{fig:result_im3} shows the failure case specific to DSIC and requirement of the TBDG model for the input floor plan image shown. The input image is a general floor plan image taken from the BRIDGE dataset, which was not included in training any model. However, DSIC yielded descriptions with details related to a plan but not relevant to the current image. Hence, with TBDG, the generated sentences describe the details of bedrooms and bathrooms, taking cues from the words. Hence it validates the robustness of the TBDG model over DSIC for a general floor plan image. 

\section{Conclusion}
\label{sec:conclusion}
In this work, we proposed models, DSIC and TBDG for generating textual description for floor plan images, which are graphical documents depicting blueprint of a building. However, being 2D line drawing images with binary pixel values, makes them different from natural images. Hence, due to lack of information at every pixel, various state of the art description generation methods for natural images do not perform well for floor plan images. Therefore, we proposed a transformer-based image to paragraph generation scheme (TBDG), which takes both image and word cues to create a paragraph. We also proposed a hierarchical recurrent neural network-based model (DSIC) to generate descriptions by directly learning features from the image, which lacks robustness in case of a general floor plan image. We evaluated the proposed model on different metrics by presenting several baselines language models for description generation and also proposing a deep learning based multi-staged pipeline to generate description from floor plan images. We trained and tested the proposed models and baselines on the BRIDGE dataset, which contains large scale floor plan images and annotations for various tasks. In future work, these models will be made more generalized to generate description for widely variable floor plan images by improving the network architecture and re-designing the method of taking word cues.

%
 \bibliographystyle{spmpsci}
%

\vspace*{2.31mm}


\vspace{-9mm}

\end{document}